# The NLP Sandbox: an efficient model-to-data system to enable federated and unbiased evaluation of clinical NLP models

Yao Yan[1,2], Thomas Yu[1], Kathleen Muenzen[3], Sijia Liu[4], Connor Boyle[1], George Koslowski[5], Jiaxin Zheng[1], Nicholas Dobbins[3], Clement Essien[1], Hongfang Liu[4], Larsson Omberg[1], Meliha Yestigen[3], Bradley Taylor[5], James A Eddy[1], Justin Guinney[3,6], Sean Mooney[3,7], Thomas Schaffter[1]

[1] Sage Bionetworks, Seattle, WA, USA

[2] Molecular Engineering & Sciences Institute, University of Washington, Seattle, WA, USA

[3] Department of Biomedical Informatics and Medical Education, University of Washington, Seattle, WA, USA

[4] Department of Artificial Intelligence and Informatics, Mayo Clinic, Rochester, MN, USA

[5] Clinical & Translational Science Institute, Medical College of Wisconsin, Milwaukee, WI, USA

[6] Tempus Lab, Chicago, IL, USA

[7] Institute for Medical Data Science, University of Washington, Seattle, WA, USA

**Abstract** (word count: 244)

**Objective**

The evaluation of natural language processing (NLP) models for clinical text de-identification relies on the availability of clinical notes, which is often restricted due to privacy concerns. The NLP Sandbox is an approach for alleviating the lack of data and evaluation frameworks for NLP models by adopting a federated, model-to-data approach. This enables unbiased federated model evaluation without the need for sharing sensitive data from multiple institutions.

**Materials and Methods**

We leveraged the Synapse collaborative framework, containerization software, and OpenAPI generator to build the NLP Sandbox (nlpsandbox.io). We evaluated two state-of-the-art NLP de-identification focused annotation models, Philter and NeuroNER, using data from three institutions. We further validated model performance using data from an external validation site.

**Results**

We demonstrated the usefulness of the NLP Sandbox through de-identification clinical model evaluation. The external developer was able to incorporate their model into the NLP Sandbox template and provide user experience feedback.

**Discussion**

We demonstrated the feasibility of using the NLP Sandbox to conduct a multi-site evaluation of clinical text de-identification models without the sharing of data. Standardized model and data schemas enable smooth model transfer and implementation. To generalize the NLP Sandbox, work is required on the part of data owners and model developers to develop suitable and standardized schemas and to adapt their data or model to fit the schemas.

**Conclusions**

The NLP Sandbox lowers the barrier to utilizing clinical data for NLP model evaluation and facilitates federated, multi-site, unbiased evaluation of NLP models.

## Introduction

**Challenges of clinical notes access**

Clinical notes contain rich information about a patient's medical history (symptoms, diagnoses, treatment plans, etc.) and may reveal important knowledge about a patient's lifestyle and disease progression. Compared to structured EHR data, clinical notes contain scattered and unstructured information, leading to difficulties in data collection, interpretation, and analysis. [1,2] Natural language processing (NLP) methods have been widely applied to clinical notes for information extraction and interpretation.[3–5] The implementation of NLP models in clinical settings often requires thorough evaluations of model performance using a large volume of clinical notes. However, due to the presence of Protected Health Information (PHI) in clinical data, access to patient data is restricted, and most models are developed and evaluated using data from limited sources, such as single institutions. This limited access to data poses challenges for the generalizability of NLP models, as data from single institutions are not always representative of other health systems.[6,7]

There are several methods that may help address the lack of clinical notes for model evaluation. Advances in synthetic clinical text generation have partially contributed to larger datasets for model development and testing.[8–12] However, due to the privacy-utility trade-off, synthetically-generated text often lacks real-world utility. Publicly available de-identified datasets like MIMIC-III [13] and i2b2 [14,15] also help alleviate the shortage of high-utility clinical data, but they are not updated regularly, and publishing de-identified sets of clinical notes poses a risk for patient re-identification.[16,17]

**Challenges of NLP model evaluation**

Challenges faced when evaluating clinical NLP models include self-assessment bias [18] and the potential lack of model generalizability. Multi-site collaborations and data sharing can help address these challenges, but it often takes time to gain administrative and governance approval, leading to research delays. Even if multi-site collaborations are approved, the data can rarely be re-used by other model developers due to privacy concerns. NLP model benchmarking frameworks like the ERASER [19], i2b2, GLUE [20], semEval [21], and Kaggle [22] challenges have attempted to address this issue by releasing standardized corpora, sets of evaluation metrics, and baseline models for comparison. However, these frameworks still use de-identified or non-sensitive datasets, and developers need to run their models on provided test datasets and upload the results to the platform for scoring, rather than uploading the actual models to the platform. As a result, the models are not automatically shared (publicly or privately with the challenge organizers) and are not immediately ready to be applied to new datasets.

**Model-to-data approach**

The model-to-data framework is a privacy-protected approach designed to lower the barrier to private data utilization.[23] Under this framework, model developers train and evaluate models using private data, but without direct access to the data. Model developers send containerized models to the data hosts which run and evaluate the models on behalf of the model developers. The only information returned to participants is model performance scores—no sensitive information leaves the data sites. This approach has already been used to support several crowdsourced DREAM challenges, including challenges focused on Electronic Health Record (EHR) data, to enable the utilization of private clinical data in the form of structured tabular data and radiology images.[24–27]

We leveraged the model-to-data approach to develop the NLP Sandbox ([nlpsandbox.io](nlpsandbox.io)) as a solution to the two problems outlined above: a lack of broadly shared, high utility clinical data, and a lack of NLP model sharing and generalizability. The NLP Sandbox is an NLP model evaluation system that enables federated evaluation and leverages knowledge and resources from multiple stakeholders. We identified three main stakeholders in the Sandbox design: (1) health institutions that contribute their datasets to the NLP Sandbox; (2) model developers who develop and submit NLP models to the sandbox environment; and (3) NLP users (including, in this case, independent validation sites) that operate external to the main NLP Sandbox infrastructure but contribute to model validation and implementation. In this project, we aim to evaluate the utility of the NLP Sandbox for comparing clinical text de-identification models. Our goal is to inspire collaboration between the stakeholders and enable the utilization of private clinical notes by the broader data science community.

## Materials and Methods

**Model evaluation system**

*Datasets*

The NLP Sandbox can be scaled to include an arbitrary number of distributed datasets. For evaluation purposes we included three datasets: (1) the 2014 i2b2 De-identification Challenge test dataset; (2) Mayo Clinic anonymized clinical notes; and (3) Medical College of Wisconsin (MCW) clinical notes (**Table 1**). The 2014 i2b2 Challenge test dataset includes 514 de-identified clinical notes. The Mayo Clinic dataset includes 148 notes with 6 note types (progress notes, plan of care, telephone encounter notes, discharge summary, consultation notes, and emergency department notes). This dataset was computationally derived by replacing name PHI detected by the MCW notes de-identification tool (https://bitbucket.org/MCW_BMI/notes-deidentification) with synthetic surrogates generated from curated common English names, and replacing date PHI with shifted dates that mimic the original format. The MCW dataset includes 433 clinical notes, all of which are progress notes.

In addition to the accessible datasets, The University of Washington (UW) is an independent validation site, meaning that NLP models can be evaluated by personnel at UW but is otherwise private. The dataset includes 956 clinical notes with 10 note types (admit, discharge, emergency department, nursing, pain management, progress, psychiatry, radiology, social work, and surgery notes), generated from 2018 to 2019. [28] The MCW and UW datasets are both fully identified. A standardized data schema is used to annotate each of the four datasets with the following five PHI types: Date, ID, Person name, Location, and Contact (Supplement 1).

*Evaluation standards*

We conduct separate evaluations of the five PHI categories in the NLP Sandbox. The PHI gold standards have been converted to our data schema in JSON format. For example, a person name PHI gold standard is annotated as {"TextPersonNameAnnotations":[{"noteId": "110-01", "start": 60, "length": 11, "text":"David Smith" }]}. We use the start position and the context of the PHI to conduct evaluation on two levels: (1) Instance level, where the evaluation is based on each complete PHI entity (e.g., "David Smith") and the model scores only if it captures both the text and location of the entity correctly, and (2) Token level, where the PHI entity is broken into tokens (e.g., "David" and "Smith" for "David Smith") and each token is evaluated independently. We report three evaluation metrics for each PHI category, both at the instance and token level: (1) Precision, (2) Recall, and (3) F1, as used in the 2014 i2b2 de-identification challenge.[29]

*Containerized NLP model*

Docker is a containerization software that uses Operating System-level virtualization to deliver applications in packages called containers. [30] Docker makes version control, model dependencies, and environment variables easier to manage and is thus conducive to model standardization and sharing. In previous model-to-data EHR DREAM Challenges,[24–27] this approach was utilized to enable the automated execution of submitted models by challenge organizers on behalf of model developers. For the NLP Sandbox, we provide a model template (Supplement 2) to reduce the barriers for developers to adapt their NLP models to the NLP Sandbox schema. This template leverages the OpenAPI generator and docker-compose, which can run multi-container applications. The model template also generates a Docker image for submission and provides step-by-step instructions for developers to build the Docker images and launch the web-based user interface (UI) for model testing.

*Evaluation workflow*

After incorporating their NLP models into the model template, model developers can submit the Docker image to Synapse, an open-source software platform developed by Sage Bionetworks that supports collaborative data science and crowdsourced challenges. [31–34] Five separate submission queues corresponding to the five PHI types are provided to developers for model performance evaluation. Once the model is submitted to one evaluation queue, the NLP Sandbox orchestrator detects the submission, automatically downloads, and runs the Docker image (**Figure 1**).

The model is initially run in the test environment, where the annotated i2b2 test dataset is stored. The NLP Sandbox orchestrator sends a request to the i2b2 data node (Supplement 3) to retrieve clinical notes and annotation gold standards. Each data node consists of a Docker container with a MongoDB-backed REST API for storing and managing clinical notes and annotation gold standards (https://github.com/nlpsandbox/data-node). The security and stability of each data node are enhanced by Nginx, [35] which is used as a reverse proxy and load balancer. Next, the NLP Sandbox orchestrator sends a request to the model and retrieves the annotation response. The NLP Sandbox orchestrator then evaluates the model annotation output against the gold standard annotation and generates scoring metrics. Finally, the orchestrator sends the scores to the leaderboard table on Synapse (**Figure 1**, Step 1). Once the model is successfully scored in the test environment, the orchestrator in the Mayo Clinic and MCW environments pulls the Docker images simultaneously to run and score the model on each dataset. The Mayo Clinic and MCW environments have similar submission infrastructure configurations as the test environment. Once model scoring is complete, the orchestrator sends the scores to the same leaderboard where the i2b2 scores are presented (**Figure 1**, Step 2).

| Data Source | | I2b2 test dataset | Medical College of Wisconsin | Mayo Clinic | University of Washington |
|---|---|---|---|---|---|
| Total no. of notes | | 514 | 433 | 148 | 956 |
| Total no. of annotations | | 10519 | 4703 | 2839 | 36843 |
| No. of annotations (%) | Date | 4980 (47.34%) | 3484 (74.08%) | 1274 (44.87%) | 21351 (56.42%) |
| | Person name | 2883 (27.41%) | 809 (17.20%) | 1565 (55.13%) | 8141 (21.51%) |
| | Contact | 218 (2.07%) | 89 (1.89%) | / | 1297 (3.43%) |
| | Id | 625 (5.94%) | 70 (1.49%) | / | 779 (2.06%) |
| | Location | 1813 (17.24%) | 251 (5.34%) | / | 6275 (16.58%) |
| No. of notes (%) | Date | 514 (100.00%) | 250 (57.74%) | 148 (100.00%) | 923 (96.55%) |
| | Person name | 508 (98.83%) | 289 (66.74%) | 148 (100.00%) | 773 (80.86%) |
| | Contact | 165 (32.10%) | 44 (10.16%) | / | 409 (42.78%) |
| | Id | 366 (71.21%) | 65 (15.01%) | / | 215 (22.49%) |
| | Location | 420 (81.71%) | 98 (22.64%) | / | 740 (77.41%) |

**Table 1**. NLP Sandbox datasets.

*Infrastructure settings*

The i2b2 and Mayo Clinic datasets are both hosted on a cloud service managed by Sage Bionetworks. The MCW environment is maintained in an on-premises server behind their institutional firewall. The UW dataset is not part of the NLP Sandbox infrastructure; this dataset is used to demonstrate how models built and evaluated in the NLP Sandbox environment can be tested on other datasets outside of the NLP Sandbox. The UW data are stored in an on-premises server behind the UW firewall. NLP Sandbox models are allotted 2 hours of total compute time, 7

GB of RAM, and 4 CPU cores during the annotation and scoring process inside the NLP Sandbox and UW environment.

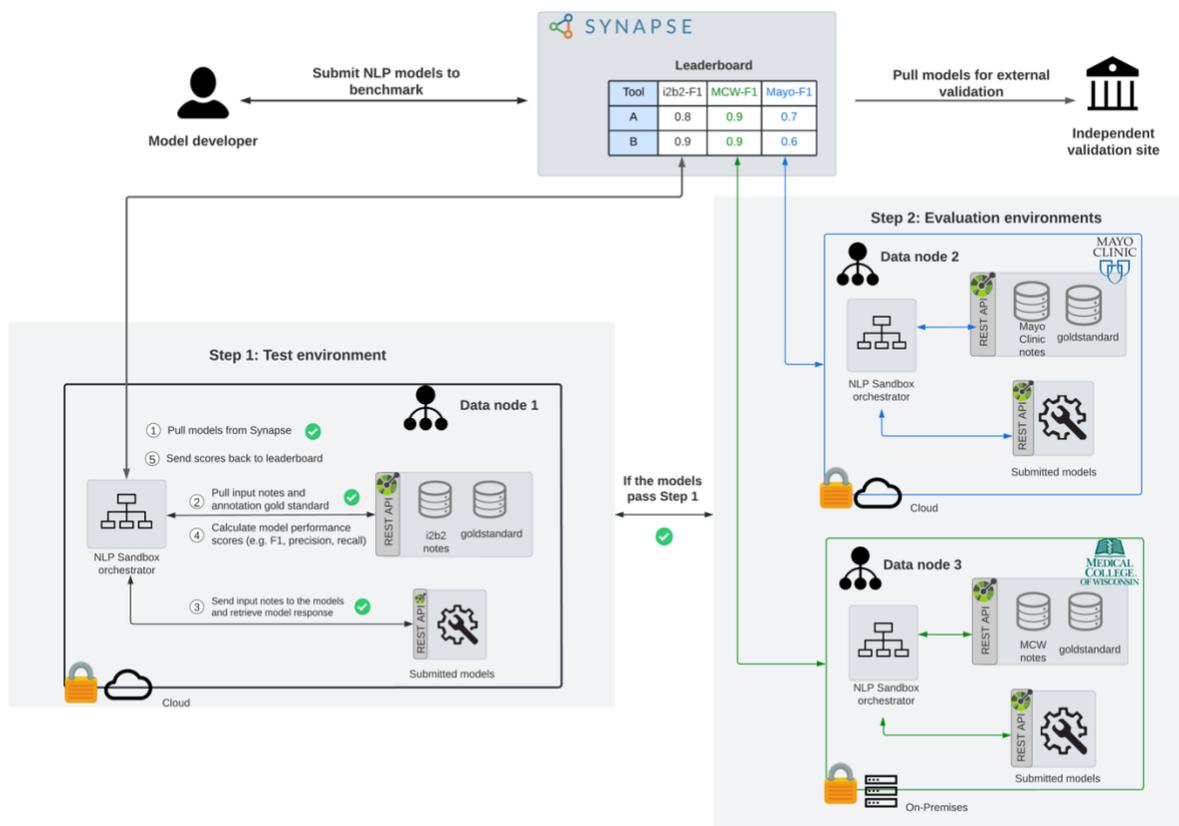

**Figure 1. NLP Sandbox evaluation workflow.** A submission made to the NLP Sandbox is first evaluated in the test environment (Step 1) on the i2b2 dataset. If the model passes Step 1, it will be sent to the Mayo Clinic and MCW environments for evaluation (Step 2). Only the scores are sent back to the NLP Sandbox leaderboard and made available to the developer.

**Experiment design**

To demonstrate the usefulness of the NLP Sandbox, we conducted three experiments: (1) User experience from a model developer, where we invited a developer of Philter [36] to adapt their Philter Python algorithm into the NLP Sandbox framework for a user experience test. The Philter developer was not involved in developing the NLP Sandbox. The purpose of this test was to assess the framework and identify limitations in an unbiased fashion; (2) Federated evaluation of an open-sourced model, where we adapted NeuroNER [37] to the NLP Sandbox and demonstrated how model benchmarking could be accomplished; and (3) Model validation using data from an independent clinical site, where we demonstrated how models submitted to the NLP Sandbox could be further applied to new datasets without the involvement of the model developer.

*External user experience test*

In the first study, we aimed to evaluate the usability of the NLP Sandbox from the perspective of an external model developer. To simulate the experience of a typical user, the developer first created an account with Synapse and became a certified user. The developer then registered to the NLP Sandbox (https://nlpsandbox.io), which is required to make submissions. There are two ways of creating an NLP Sandbox submission: 1) using the NLP Sandbox model template that we provide and 2) generating a "tool stub" for any of the programming languages-frameworks supported by the OpenAPI Generator using one of the tool specifications provided by the NLP Sandbox. The developer was asked to clone the NLP Sandbox model template from GitHub.(Supplement 2) The template contained separate modules corresponding to each PHI category. In each module, the data structure and format of input notes and annotation outputs were specified. To test whether the model ran, the developer created a Docker image following instructions in the model template GitHub page and tested the model locally through a local Swagger UI webpage that was automatically generated by OpenAPI Generator.[38]

To submit the model to the NLP Sandbox, the developer tagged the Docker image with a unique version number and pushed the tagged image to their private Docker repository on Synapse. The developer then submitted the tagged image to the five different submission queues for each PHI category, where each queue represented an annotation task. For a given task, the developer received detailed information about the performance of their model on each dataset, including the F1 score, precision, and recall. If the submission failed in the test environment, the developer received an email notification with a link to the error logs generated from the i2b2 test environment. For more detailed troubleshooting, the developer worked with an NLP Sandbox developer to review the model's performance on the i2b2 dataset. Because the i2b2 dataset is public and fully de-identified, an NLP Sandbox developer returned false positive and false negative annotation results to the model developer for further model improvement. To ensure privacy protection and unbiased evaluation, no log files, false positives, or false negatives from the MCW or Mayo Clinic datasets were provided to the Philter developer. The developer went through several iterations of this process to improve the submission. Finally, the Philter developer was asked to provide an assessment of the model template adaptation and submission process.

*Federated evaluation for an open-sourced model*

In the second study, we wanted to explore the benchmarking that is feasible to conduct within the current NLP Sandbox framework. The NLP Sandbox currently does not make any identified datasets available for model training to model developers. However, if the developer has access to a private dataset, they can train their model using this dataset and submit the pre-trained model to the NLP Sandbox for model evaluation. We used NeuroNER, a neural network-based de-identification system, to demonstrate how a pre-trained machine learning model could be adapted to the NLP Sandbox.

To conduct the experiment, we first incorporated the NeuroNER package and i2b2 pre-trained model into the NLP Sandbox model template. Then, we submitted the Docker image to Synapse and evaluated NeuroNER using the i2b2 test, Mayo Clinic, and MCW datasets. Following the evaluation process, the model performance scores for the three sites were returned to the Synapse leaderboard.

*Model validation using data from an independent site*

To mimic the further implementation of models submitted to the NLP Sandbox at external sites, we used a private dataset from UW and an on-premises server behind the UW firewall to conduct an independent validation of submitted models (**Figure 2**). We first translated the UW data into the NLP Sandbox data schema and conducted a thorough data quality check to ensure that there were no annotation duplicates and that the start and end position of each annotation in the gold standard matched those in the original notes. We launched the data node service on the UW server to host the private clinical notes. We then used Docker commands to pull the submitted NeuroNER and Philter Docker images from Synapse and run them in the UW environment. The NLP Sandbox provides a Python package called nlpsandbox-client (https://pypi.org/project/nlpsandbox-client) that enables users to pull clinical notes from a data node (https://github.com/nlpsandbox/data-node), process them using the submitted models, and evaluate model performance by generating F1 score, precision, and recall for each PHI annotation task registered in the NLP Sandbox.

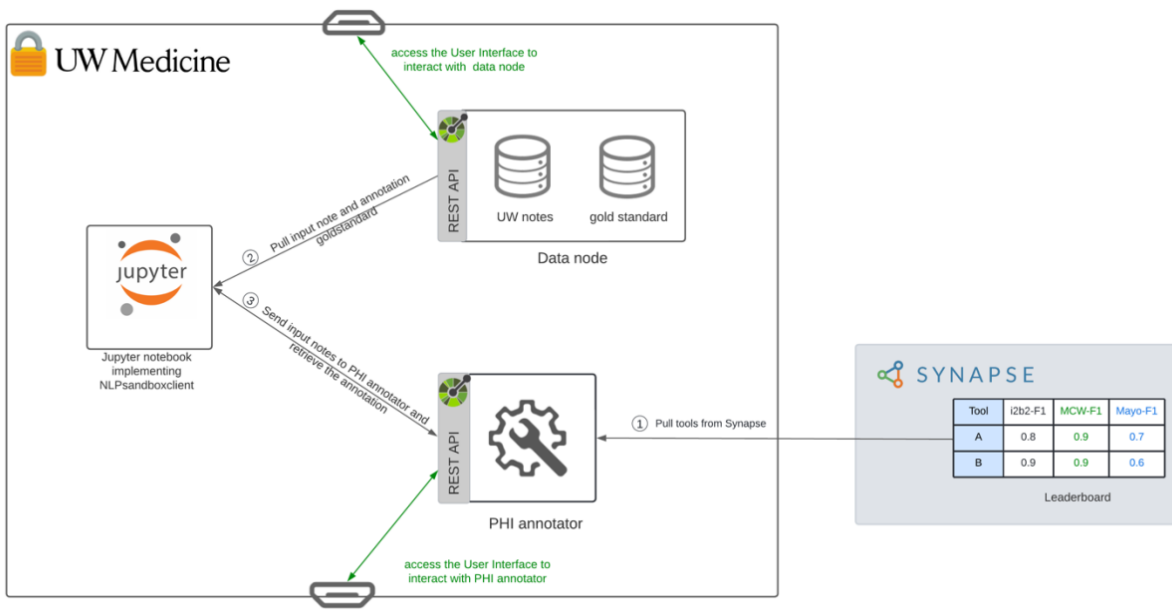

**Figure 2. Infrastructure of the independent validation site (UW) environment.** This server hosts the UW clinical notes and is behind a UW firewall. For each model validation test, we pulled the Docker image of an NLP Sandbox PHI annotator model submission from Synapse and deployed it inside the UW environment. A Python package, nlpsandbox-client is imported into a Jupyter notebook to pull notes from the data node, interact with the NLP Sandbox PHI annotator, and conduct the evaluation. Both the PHI

annotator and data node provide a UI webpage that can be used to interact with them, for testing or demonstration purposes.

## Results

**User experience from a model developer**

The developer reported no difficulties in setting up a user account with Synapse, but they mentioned that cloning and modifying the de-identification model GitHub template required both command-line experience and knowledge of the Git version control tool. The developer worked with an NLP Sandbox developer to complete both the initial model integration and Docker image submission successfully. During model integration, the developer reported that the UI webpage was helpful for identifying code integration errors and assessing whether each PHI annotator was functioning as expected. Although detailed stack trace messages did not appear in the UI, high-level error codes were displayed and helped guide debugging efforts. The developer was also able to generate print statements to retrieve more detailed debugging information.

In the original version of Philter, de-identification rules were applied sequentially as defined in a JSON configuration file. The Philter package utilizes a mix of rules including regular expressions, exclude regular expressions, include lists, exclude lists, and advanced pattern matching using previously identified PHI. To package the rule-based Philter algorithm as an NLP Sandbox PHI annotator, the developer initially incorporated only exclude regular expressions and lists directly into the Python PHI annotation files. Because the NLP Sandbox evaluates the model based on its performance in each PHI task independently, the developer was required to apply different de-identification rules in separate annotation modules for each PHI type. This resulted in low precision and recall scores compared to previously reported scores on the 2014 i2b2 test corpus. Previously reported Philter recall scores on the 2014 i2b2 corpus were above 0.96, and precision scores were above 0.78 across all PHI categories. In contrast, recall scores evaluated within the NLP Sandbox environment ranged from 0.74-0.99 on the 2014 i2b2 corpus, and precision ranged from 0.37-0.79 for different PHI categories.

To rescue recall and precision as much as possible, the developer imported a modified version of the Philter package into the NLP Sandbox model template to leverage additional de-identification rules. However, these changes did not significantly improve performance, indicating that poor performance was not necessarily caused by unused algorithmic rules, but was instead caused by differences in evaluation standards. In previously published evaluations of Philter, the evaluation was category-agnostic, and overlapping identification of PHI by different rules did not impact precision as long as the PHI was obscured. In the current NLP Sandbox evaluation, the evaluation is conducted separately for each PHI task. If a token is categorized as a PHI type different from the gold standard, this could negatively impact performance.

Overall, the developer recommended that clearer and more thorough documentation of the development process be posted to the GitHub README, including documentation of how to incorporate algorithms wholesale into the model schema. Additionally, returning i2b2 false positives and false negatives to developers may help them diagnose issues with model integration.

**Model performance**

After several iterations of model adaptation and submission, the Philter developer received scores for all five PHI tasks (**Table 2**). Philter achieved the highest F1 score, precision, and recall on date annotation across datasets, compared to other PHI categories, and the lowest performances on ID and location. Token-level evaluation scores exceeded instance-level evaluation scores in most tasks, with the exception of ID annotation on the MCW and UW dataset.

NeuroNER also achieved the highest scores for date annotation and the lowest scores for ID annotation (**Table 3**). Overall, NeuroNER outperformed Philter. The performance of NeuroNER on the i2b2 test dataset was significantly higher than other datasets on all PHI categories except date.

| Dataset | Evaluation standard | PHI Category (P/R/F1) | | | | |
|---|---|---|---|---|---|---|
| | | Date | Person Name | ID | Contact | Location |
| i2b2 | instance | 0.77/0.88/0.82 | 0.29/0.68/0.41 | 0.34/0.90/0.49 | 0.31/0.99/0.47 | 0.27/0.23/0.25 |
| | token | 0.79/0.89/0.84 | 0.51/0.86/0.64 | 0.37/0.94/0.53 | 0.43/0.99/0.60 | 0.52/0.74/0.61 |
| Mayo Clinic | instance | 0.95/0.99/0.97 | 0.25/0.55/0.34 | / | / | / |
| | token | 0.95/0.99/0.97 | 0.71/1.00/0.83 | / | / | / |
| MCW | instance | 0.76/0.94/0.84 | 0.19/0.63/0.29 | 0.13/0.90/0.23 | 0.25/0.85/0.39 | 0.09/0.14/0.11 |
| | token | 0.77/0.94/0.85 | 0.41/0.86/0.56 | 0.10/0.89/0.18 | 0.30/0.87/0.45 | 0.28/0.51/0.36 |
| UW | instance | 0.85/0.91/0.88 | 0.19/0.62/0.29 | 0.22/0.83/0.35 | 0.38/0.64/0.48 | 0.15/0.14/0.14 |
| | token | 0.87/0.92/0.89 | 0.38/0.84/0.52 | 0.21/0.79/0.33 | 0.39/0.81/0.53 | 0.27/0.43/0.33 |

**Table 2. Philter performance on all NLP Sandbox test datasets.** Evaluation metrics in each cell are represented as precision/recall/F1. The darker red colors correspond to higher F1 scores and the darker blue colors correspond to lower F1 scores.

| Dataset | Evaluation standard | PHI Category (P/R/F1) | | | | |
|---|---|---|---|---|---|---|
| | | Date | Person Name | ID | Contact | Location |
| i2b2 | instance | 0.92/0.91/0.91 | 0.91/0.88/0.89 | 0.67/0.65/0.66 | 0.86/0.93/0.89 | 0.86/0.79/0.82 |
| | token | 0.93/0.92/0.92 | 0.96/0.93/0.94 | 0.65/0.69/0.67 | 0.91/0.95/0.93 | 0.96/0.86/0.91 |
| Mayo Clinic | instance | 0.92/0.99/0.95 | 0.24/0.20/0.22 | / | / | / |
| | token | 0.91/1.00/0.95 | 0.51/0.76/0.61 | / | / | / |
| MCW | instance | 0.89/0.95/0.92 | 0.76/0.80/0.78 | 0.51/0.70/0.59 | 0.42/0.76/0.54 | 0.37/0.36/0.36 |
| | token | 0.89/0.95/0.92 | 0.87/0.85/0.86 | 0.45/0.67/0.54 | 0.43/0.75/0.55 | 0.60/0.40/0.48 |
| UW | instance | 0.92/0.93/0.92 | 0.79/0.68/0.73 | 0.40/0.42/0.41 | 0.60/0.76/0.67 | 0.62/0.47/0.53 |
| | token | 0.93/0.94/0.93 | 0.86/0/77/0.81 | 0.36/0.40/0.38 | 0.66/0.76/0.71 | 0.68/0.48/0.56 |

**Table 3. NeuroNER performance on all NLP Sandbox test datasets.** Evaluation metrics in each cell are represented as F1/precision/recall. The darker red colors correspond to higher F1 scores and the darker blue colors correspond to lower F1 scores.

## Discussion

In this pilot study, we achieved our goal of evaluating NLP models for PHI annotation using multiple public and private datasets. We demonstrated how health institutions, model developers, and independent validation sites all can make use of the NLP Sandbox to accomplish their tasks. The successful operation of the NLP Sandbox requires (1) a standardized but extensible model and data schema; (2) the participation of model developers; (3) the involvement of health institutions contributing their private data for evaluation; and (4) the engagement of external sites to extend the usage of submitted models for validation and implementation outside of the NLP Sandbox environment.

**A federated evaluation framework enabling secure data utilization without granting data access**

NLP applications are growing in popularity in the healthcare industry. It is widely recognized that sharing clinical notes more broadly with the data science community can improve the development of Artificial Intelligence (AI) strategies to address clinically-relevant questions and improve healthcare quality. However, bounded by privacy concerns, health institutions are cautious and often progress slowly when it comes to sharing clinical data with both internal and external researchers. The NLP Sandbox offers a secure way for health institutions to open up their data for model evaluation without granting direct access to sensitive data. Under the NLP Sandbox model-to-data framework, health institutions have full control over how their data is used in the NLP Sandbox environment. They can choose to deposit their data in any secure computational environment (e.g., on-premises server, AWS, Azure, etc.) behind their institution's firewalls to make their datasets available in NLP Sandbox data nodes. Only the models themselves, rather than the model developers, can access the data for evaluation. In this way, no sensitive patient information leaves the health institution.

**Federated and unbiased evaluation of NLP models in a ready-to-go setting**

When pre-trained using the 2014 i2b2 training dataset, NeuroNER achieved the highest performance on test data from the same source while the performance did not generalize as well on datasets from different sources. Failure to test model generalizability and self-assessment bias are two main criticisms facing NLP model developers. However, they can be avoided if model developers use an integrated development and testing environment like the NLP Sandbox. Our model-to-data framework lowers the barrier of access to clinical notes and can accelerate the development and evaluation of NLP models. Developers do not need to wait for months to get approval to access and implement their models on identified clinical notes from other institutions, and can quickly retrieve multi-site performance metrics that can be used to inform further model development.

We also made efforts to lower the barrier for developers to adapt their models for submission to the NLP Sandbox. Our model template offers step-by-step instructions for developers to build a Docker image, test the incorporated model in an interactive UI, and submit the model for evaluation. In the first case study, we demonstrated that the NLP Sandbox can be used to iteratively return feedback to developers who wish to improve model performance.

**Model and data standardization for future application**

The NLP Sandbox utilizes a standardized data schema and applies quality checks on data converted to the schema. After converting their data to the NLP Sandbox schema, data owners external to the NLP Sandbox can pull the models submitted to the NLP Sandbox from Synapse and seamlessly test the models on their data without exposing their data to the developers. This allows future users to implement models submitted to the NLP Sandbox and allows for continuous benchmarking of existing NLP Sandbox submissions on newly incorporated data nodes. Our schema is also flexible and can support tasks beyond PHI annotation, as demonstrated in a pilot test using the NLP Sandbox schema to support COVID symptom annotation (Supplement 4).

**Lesson learned and limitations**

We identified several aspects of the NLP Sandbox with room for improvement:

**(1) Model template.** The external model developer reported difficulties with incorporating model scripts into the NLP Sandbox. We already updated documentation but the Sandbox could benefit from additional detailed instructions and video tutorials to help developers incorporate their models into the provided template. In particular, instructions for wrapping models as packages, importing custom packages into the template, and making annotation outputs compatible with the NLP Sandbox data schema are helpful.

**(2) Data schema.** Our current data schema cannot suit all evaluation needs, as shown in the Philter case, and the involvement of the model developing community is needed for defining standardized schemas. Currently, NLP Sandbox implemented the schema used for the 2014 i2b2 PHI De-identification Challenge. For models like NeuroNER that were designed with i2b2 schemas in mind, they naturally fit and transitioned smoothly into the NLP Sandbox. On the other hand, Philter was not developed with the i2b2 data schema in mind, and instead prioritized identifying PHI regardless of the type of PHI and tolerated duplicated annotations to maximize recall. As a result, Philter's performance suffered in the NLP Sandbox. For example, Philter was penalized for capturing "David", "Smith," and "David Smith" as separate instances of PHI, and for identifying "David Smith" as both a location and a name. The flexibility of NLP Sandbox makes it easy to add separate evaluation queues that can be used to answer different questions, such as PHI category-agnostic performance in the future. Additionally, our decision to implement a common data schema enables data standardization but puts the burden on data owners to adapt their data to our schema. One future direction would be to allow data providers to specify the data schema they used to annotate the dataset and then tailor the NLP Sandbox evaluation to the unique schema of each dataset. Alternatively, we can automate the generation of multiple data schemas for each dataset and users could compare model performance across different schemas.

**(3) Model training.** While model training is not currently enabled in the NLP Sandbox, rule-based and pre-trained NLP models are accepted for evaluation. We observed from the NeuroNER experiment that training and testing models using data from the same site can lead to better performance, especially for PHI categories like location and ID where each site has its format and specification. However, pre-trained models often carry fragments of the data that they are trained on. To achieve model training inside the NLP Sandbox, additional data use agreements and security measures must be implemented to ensure that models trained on sensitive datasets do not leak PHI in other data nodes.

**(4) Data quality.** While we applied data quality checks on the i2b2, MCW, Mayo Clinic, and UW datasets, it is possible that the gold standard annotations used for evaluation are not 100% correct or are not representative of authentic clinical notes. For example, we observed some quality issues in the synthetically de-identified Mayo Clinic data (e.g., unusually long patient names) leading to low annotation performance for NeuroNER.

## Conclusions

In this study, we demonstrated that the NLP Sandbox, as a model-to-data evaluation system, enables the privacy-protected utilization of clinical notes and unbiased federated evaluation of NLP models applied to clinical notes. Model developers can receive an informative model evaluation for multiple datasets that can be used to improve model performance and generalizability. The standardization of data and model schemas in the NLP Sandbox enables smooth implementation in the production setting and thus increases the potential for the further application of the models submitted to the NLP Sandbox.

**Funding**

This study was supported by the Clinical and Translational Science Awards Program National Center for Data to Health funding by the National Center for Advancing Translational Sciences at the National Institutes of Health, grant U24TR002306 [YY, TY, TS, JG, JE, LO]; UL1TR001436[GK, BT], and grant U01TR001062 [SL, HL].

**Conflicts of interest**

The authors have no competing conflicts of interest.

# Supplement

## 1. Data schema

We establish a data schema for clinical notes and corresponding annotation goldstandards. (https://github.com/nlpsandbox/nlpsandbox-schemas/tree/main/openapi/commons/components/schemas, v1.2.0) Once the data is converted to a JSON file compliant with the schema, it can be pushed and deposited into an instance of the NLP Sandbox Data Node (https://github.com/nlpsandbox/data-node).

## 2. Model template

### *Incorporate NLP models to the template*

We provide a model template (https://github.com/nlpsandbox/phi-annotator-example, v1.2.1) for developers to convert their models for submission to the NLP Sandbox. Model developers need to incorporate their model into the 5 modules (controller files) (https://github.com/nlpsandbox/phi-annotator-example/tree/main/server/openapi_server/controllers) for the 5 PHI categories and update the requirement.txt (https://github.com/nlpsandbox/data-node/blob/main/server/requirements.txt) for adding new packages they want to import. Once it's done, developers can follow the step-by-step instructions in the README to build a Docker image for submission and a UI webpage for model testing.

### *Submitting the Docker images to the NLP Sandbox*

We provide detailed instructions (https://www.synapse.org/#!Synapse:syn22277123/wiki/609136) for how to make submissions to the NLP Sandbox.

### *Accessing the UI for model testing*

The UI webpage generated by the OpenAPI generator is accessible through the link in the README. Here we use date annotation to demonstrate how the UI can be used for model testing.

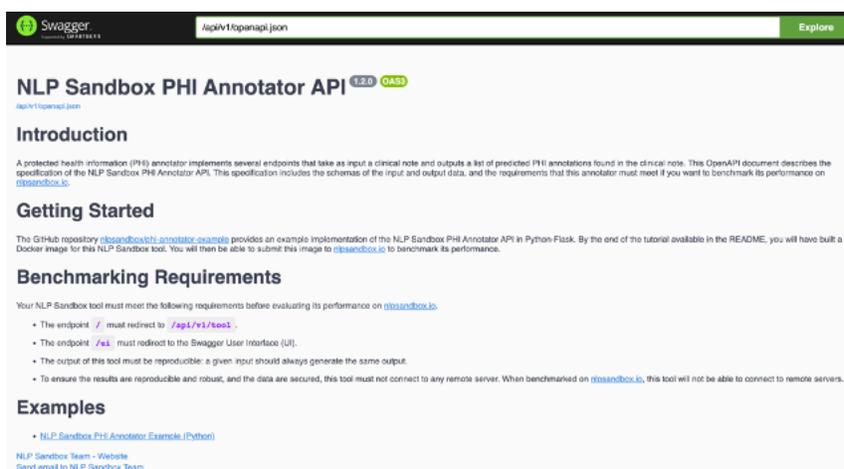

## TextDateAnnotation — Operations about text date annotations

**POST** `/textDateAnnotations` Annotate dates in a clinical note

Return the date annotations found in a clinical note

**Parameters**

No parameters

Request body — application/json

```json
{
  "note": {
    "identifier": "awesome-note",
    "patientId": "awesome-patient",
    "text": "On 12/26/2020, Ms. Chloe Price met with Dr. Prescott in Seattle.",
    "type": "loinc:LP29684-5"
  }
}
```

[Execute] [Clear]

**Text for testing can be modified to any sentence. Clicking on the "Execute" button in blue allows the PHI annotator to run on the test sentence and generate date annotations.**

**Responses**

Curl

```
curl -X 'POST' \
  'http://localhost:8081/api/v1/textDateAnnotations' \
  -H 'accept: application/json' \
  -H 'Content-Type: application/json' \
  -d '{
  "note": {
    "identifier": "awesome-note",
    "patientId": "awesome-patient",
    "text": "On 12/26/2020, Ms. Chloe Price met with Dr. Prescott in Seattle.",
    "type": "loinc:LP29684-5"
  }
}'
```

Request URL

`http://localhost:8081/api/v1/textDateAnnotations`

Server response

| Code | Details |
|------|---------|
| 200  | Response body |

```json
{
  "textDateAnnotations": [
    {
      "confidence": 95.5,
      "dateFormat": "MM/DD/YYYY",
      "length": 10,
      "start": 3,
      "text": "12/26/2020"
    },
    {
      "confidence": 95.5,
      "dateFormat": "YYYY",
      "length": 4,
      "start": 9,
      "text": "2020"
    }
  ]
}
```

*Dockerized model examples*

NeuroNER:
- Github https://github.com/yy6linda/phi-annotator-neuroner
- Docker image: docker.synapse.org/syn22277123/phi-annotator-neuroner-all-types-yao@sha256:61b0f50fb453a0ec6d5a0a2dfb0573895b96b9ba766aa32bedbc8d9d2ca26c21

Philter
- Github: https://github.com/kmuenzen/nlpsandbox_philter_allphi
- Docker image: docker.synapse.org/syn26450115/nlpsandbox_philter_allphi@sha256:b394dc0217bbd015e55971eafb5e6df9d743e7d536960adb55c70078164aad5f

### 3. NLP Sandbox Data Node

We launch a data node service to host the clinical notes and annotation gold standards (https://github.com/nlpsandbox/data-node, v1.2.1). The data node is a REST API deployed using a Docker container along with a MongoDB database instance. Nginx is used as a reverse proxy and load balancer to enhance the security and stability of the data node. We implement OpenAPI generator for the data node service, which enables the creation of a User Interface(UI) web page for users to pull, deposit, and view clinical notes (Supplement Figure 1).

Supplement Figure 1. A snapshot of the data node UI webpage

## 4. COVID symptom annotation

The NLP Sandbox also enables the evaluation of COVID symptom annotators, which we launched in part to demonstrate that the NLP Sandbox can be used to benchmark non-PHI annotation tools. The goal of this task is to annotate COVID symptoms in clinical notes. We used synthetic data from Mayo Clinic for this task. The data are formatted using the NLP Sandbox COVID annotation data schema (https://github.com/nlpsandbox/nlpsandbox-schemas/blob/main/openapi/commons/components/schemas/TextCovidSymptomAnnotation.yaml) The data include 94 notes and 533 annotations.

We provide a baseline model for this task (https://github.com/nlpsandbox/covid-symptom-annotator-example, v1.2.0). The baseline model achieves 0.71/0.79/0.64(F1/precision/recall) in instance-level evaluation and 0.70/0.89/0.58 in token-level evaluation.